\documentclass[11pt,a4paper]{article}
\usepackage[hyperref]{acl2021}
\usepackage{times}
\usepackage{latexsym}

\usepackage{microtype}

\usepackage[]{todonotes}
\usepackage{colortbl}
\usepackage{graphicx,multirow}

\aclfinalcopy 

\title{IMS' Systems for the IWSLT 2021 Low-Resource Speech Translation Task}

\author{Pavel Denisov, Manuel Mager, Ngoc Thang Vu \\
  Institute for Natural Language Processing, University of Stuttgart \\
  \texttt{\{pavel.denisov,manuel.mager,thangvu\}@ims.uni-stuttgart.de}}

\date{}

\begin{document}
\maketitle
\begin{abstract}
This paper describes the submission to
the IWSLT 2021 Low-Resource Speech Translation Shared Task by IMS team.
We utilize state-of-the-art models
combined with several data augmentation, multi-task and transfer
learning approaches for the automatic speech recognition (ASR) and machine translation (MT) steps of
our cascaded system. 
Moreover, we also explore the feasibility of a full end-to-end speech translation (ST)
model in the case of very constrained amount of ground truth labeled data.
Our best system achieves the best performance
among all submitted systems for Congolese Swahili to English and French
with BLEU scores 7.7 and 13.7 respectively,
and the second best result for Coastal Swahili to English
with BLEU score 14.9.
\end{abstract}

\section{Introduction}

We participate in the low-resource speech translation task
of IWSLT 2021. This task is organized for the first time, and
it focuses on three speech translation directions this year:
Coastal Swahili to English (\texttt{swa}\textrightarrow \texttt{eng}),
Congolese Swahili to French (\texttt{swc}\textrightarrow \texttt{fra}) and
Congolese Swahili to English (\texttt{swc}\textrightarrow \texttt{eng}).
Working on under-represented and low-resource languages is 
of special relevance for the inclusion into technologies of
big parts of the world population. The Masakhane initiative \cite{nekoto-etal-2020-participatory}
has opened the doors for large scale participatory research on
languages of the African continent, to which Swahili belongs to. 
Our Speech-to-Text translation systems aim to contribute to
this global effort.

A common problem for these languages is the small amount of data.
This is also true for the language pairs of the shared task: the
provided data contains a small amount of translated speech samples for each pair,
but the participants are allowed to use additional data and pre-trained models
for the sub-tasks of ASR and MT.
We utilize most of the suggested additional data resources
to train and tune sequence-to-sequence ASR and MT components.
Our primary submission is the cascaded system built of Conformer end-to-end
ASR model and Transformer MT model.
Our contrastive system is end-to-end ST system
utilizing parameters transfer from the Encoder
part of ASR model and the full MT model.

Both ASR and MT components of the cascaded system initially
yield good results on their own, but the discrepancy between
language formats (spoken vs. written) in ASR and MT corpora
causes degradation by 47\% in resulting scores. 
To adapt the MT system to the output of the ASR, we transform the Swahili source data to output similar to one of an ASR system. 
To further increase the performance of our MT system, we leverage 
both source formats (original Swahili text and simulated ASR output Swahili) 
into a multi-task framework. This approach improves our results by $17\%$, mostly for the English target language. 
Our system outperforms the next best system on \texttt{swc}\textrightarrow \texttt{fra} 
by $4.4$ BLEU points, but got outperformed by $10.4$ BLEU for \texttt{swa}\textrightarrow \texttt{eng},
being the second-best team. Our team was the only participating for 
\texttt{swc}\textrightarrow \texttt{eng} language pair with a score of $7.7$ BLEU.
The results of end-to-end system consistently appear to be about twice
worse compared to the pipeline approach.

\section{ASR}
\label{sec:asr}

\subsection{Data}
\label{subsec:asrdata}
Table \ref{tab:data:asr} summarizes the datasets used to develop our ASR system.
The training data comprises of the shared task training data,
Gamayun Swahili speech samples\footnote{\url{https://gamayun.translatorswb.org/data/}}
and the training subsets of
ALFFA dataset \citep{gelas2012developments}
and IARPA Babel Swahili Language Pack \cite{andresen2017iarpa}.
The validation data comprises of 869 randomly sampled
utterances from the shared task training data and the
testing subset of ALFFA dataset.
The testing data is the shared task's validation data.
All audio is converted to 16 kHz sampling rate.
Applied data augmentation methods are speed perturbation with the factors
of 0.9, 1.0 and 1.1, as well as SpecAugment \citep{park2019specaugment}.
Transcriptions of the shared task data and Gamayun Swahili speech samples dataset
are converted from written to spoken language similarly to \citet{bahar2020start},
namely all numbers are converted to
words\footnote{Using \url{http://www.bantu-languages.com/en/tools/swahili\_numbers.html}},
punctuation is removed and letters are converted to lower case.
External LM is trained on the combination of transcriptions of the ASR training data
and LM training data from ALFFA dataset. The validation data for the external LM
contains only transcriptions of the ASR validation data.

\begin{table}[ht]
  \centering
  \small
  \setlength{\tabcolsep}{2pt}
  \begin{tabular*}{1.0\columnwidth}{l|r|r|r|r|r|r}
    \noalign{\hrule height 1pt}
    Dataset & \multicolumn{2}{|c|}{Training} & \multicolumn{2}{|c|}{Validation} & \multicolumn{2}{|c}{Testing} \\
    \cline{2-7}
                        & \multicolumn{1}{|c|}{Utt.} & \multicolumn{1}{|c|}{Hours} & \multicolumn{1}{|c|}{Utt.}  & \multicolumn{1}{|c|}{Hours}  & \multicolumn{1}{|c|}{Utt.} & \multicolumn{1}{|c}{Hours} \\
    \hline
    IWSLT'21 \texttt{swa}& 4,162  & 5.3   & 434    & 0.5   & 868   & 3.7  \\
    IWSLT'21 \texttt{swc}& 4,565  & 11.1  & 435    & 1.0   & 868   & 4.2  \\
    Gamayun              & 4,256  & 5.4   & -      & -     & -     & -    \\
    IARPA Babel          & 21,891 & 28.8  & -      & -     & -     & -    \\
    ALFFA                & 9,854  & 9.5   & 1,991  & 1.8   & -     & -    \\
    \hline
    Total                & 44,728 & 60.3  & 2,860  & 3.4   & 1,736 & 7.9  \\
    \hline
    \noalign{\hrule height 1pt}
  \end{tabular*}
  \caption{
    \label{tab:data:asr} Datasets used for the ASR system.
  }
\end{table}

\subsection{Model}
\label{subset:asrmodel}

The ASR system is based on end-to-end Conformer ASR \citep{gulati2020conformer}
and its ESPnet implementation \citep{guo2020recent}.
Following the latest LibriSpeech recipe \citep{kamo2021espnet2},
our model has 12 Conformer blocks in Encoder and 6 Transformer blocks
in Decoder with 8 heads and attention dimension of 512.
The input features are 80 dimensional log Mel filterbanks.
The output units are 100 byte-pair-encoding (BPE) tokens \citep{sennrich2016neural}.
The warm-up learning rate strategy \citep{vaswani2017attention} is used, while the learning rate
coefficient is set to 0.005 and the number of warm-up steps is set to 10000.
The model is optimized to jointly minimize cross-entropy and connectionist temporal
classification (CTC) \citep{graves2006connectionist} loss functions,
both with the coefficient of 0.5.
The training is performed for 35 epochs on 2 GPUs with the total
batch size of 20M bins and gradient accumulation over each 2 steps.
After that, 10 checkpoints with the best validation accuracy are averaged for the decoding.
The decoding is performed using beam search with the beam size of 8
on the combination of Decoder attention
and CTC prefix scores \citep{kim2017joint} also with the coefficients of 0.5 for both.
In addition to that, external BPE token-level language model (LM)
is used during the decoding in the final ASR system.
The external LM has 16 Transformer blocks with 8 heads and attention dimension of 512.
It is trained for 30 epochs on 4 GPUs
with the total batch size of 5M bins, the
learning rate coefficient 0.001 and 25000 warm-up steps.
Single checkpoint having the best validation perplexity is used for the decoding.

\subsection{Pre-trained models}
\label{subsec:asrpretrained}

In addition to training from scratch,
we attempt to fine-tune several pre-trained speech models.
These models include ESPnet2 Conformer ASR models from the LibriSpeech \citep{panayotov2015librispeech},
SPGISpeech \citep{o2021spgispeech} and
Russian Open STT\footnote{\url{https://github.com/snakers4/open_stt}}
recipes, as well as wav2vec 2.0 \citep{baevski2020wav2vec} based models XLSR-53 \citep{conneau2020unsupervised}
and VoxPopuli \citep{wang2021voxpopuli}.

\subsection{Results}
\label{subsec:asrresults}
Table \ref{tab:results:asr} summarizes the explored ASR settings and the results
on the shared task validation data. CTC weight 0.5 is selected in order
to minimize the gap between ASR accuracy on the two Swahili languages.
Evaluation of pre-trained English ASR models expectedly shows that
SPGISpeech model results in better WER, likely because of the larger
amount of training data or more diverse accent representation in this corpus
compared to LibriSpeech. Surprisingly, pre-trained Russian Open STT
model yields even better results than SPGISpeech model, even if the amount
of the training data for them is quite similar (about 5000 hours).
Since Swahili language is not closely related to English or Russian,
we attribute better results of Russian Open STT model either to
the larger amount of acoustic conditions and speaking styles in Russian Open STT corpus,
or to more similar output vocabulary in the model: both Russian and Swahili models
use 100 subword units, while English models use 5000 units.
Validation accuracy of wav2vec 2.0 models does not look promising
in our experiments and we do not include their decoding results to the table.
Freezing the first Encoder layer of Russian Open STT model
during training on Swahili data
gives us consistent improvement on both testing datasets,
but freezing more layers does not appear to be beneficial.
Interestingly enough, external LM also improves results on both
Coastal and Congolese Swahili, however the best LM weights
differ between languages, and we conclude to keep them separate in
the final system.

\begin{table}[ht]
  \centering
  \small
  \setlength{\tabcolsep}{3.5pt}
  \begin{tabular*}{1.0\columnwidth}{l|l|l|l|l}
    \noalign{\hrule height 1pt}
    \# & System & \texttt{swa} & \texttt{swc} & Avg. \\
    \hline
    1. & CTC weight 0.3 & 25.9 & 26.5 & 26.2 \\
    2. & CTC weight 0.4 & 25.8 & 24.4 & 25.1 \\
    3. & CTC weight 0.5 & 25.2 & 25.0 & \textbf{25.1} \\
    4. & CTC weight 0.6 & 25.4 & 25.0 & 25.2 \\
    5. & CTC weight 0.7 & 26.4 & 24.9 & 25.7 \\
    \hline
    6. & \#3, pre-trained LibriSpeech & 22.4 & 25.4 & 23.9 \\
    7. & \#3, pre-trained SPGISpeech & 20.8 & 22.9 & 21.9 \\
    8. & \#3, pre-trained Russian Open STT & 21.4 & 20.8 & \textbf{21.1} \\
    \hline
    9. & \#8, freeze Encoder layers \#1--4 & 20.3 & 21.1 & 20.7 \\
    10. & \#8, freeze Encoder layers \#1--2 & 21.9 & 21.4 & 21.7 \\
    11. & \#8, freeze Encoder layer \#1 & 17.8 & 19.7 & \textbf{18.8} \\
    \hline
    12. & \#11, average 9 checkpoints & 17.7 & 19.7 & 18.7 \\
    13. & \#11, average 8 checkpoints & 17.7 & 19.5 & \textbf{18.6} \\
    14. & \#11, average 7 checkpoints & 17.8 & 19.6 & 18.7 \\
    15. & \#11, average 6 checkpoints & 17.7 & 19.5 & 18.6 \\
    16. & \#11, average 5 checkpoints & 17.9 & 19.6 & 18.8 \\
    \hline
    17. & \#13, external LM weight 0.2 & 15.1 & 18.4 & 16.8 \\
    18. & \#13, external LM weight 0.3 & 14.5 & \textbf{18.3} & 16.4 \\
    19. & \#13, external LM weight 0.4 & 14.0 & 18.5 & 16.3 \\
    20. & \#13, external LM weight 0.5 & 13.6 & 18.7 & 16.2 \\
    21. & \#13, external LM weight 0.6 & \textbf{13.5} & 19.1 & 16.3 \\
    22. & \#13, external LM weight 0.7 & 13.8 & 19.9 & 16.9 \\
    \noalign{\hrule height 1pt}
  \end{tabular*}
  \caption{
    \label{tab:results:asr} ASR results (WER, \%) on the shared task validation data.
    Bold numbers correspond to the selected configuration for the final system
    (the external LM weights are language-specific).
  }
\end{table}

\section{MT}
\label{sec:mt}

\subsection{Data}
\label{subsec:mtdata}

Table \ref{tab:data:mt} summarizes the datasets used to train our MT systems.
The training data comprises of the shared task training data,
Gamayun kit\footnote{\url{https://gamayun.translatorswb.org/data/}}
(English – Swahili and
Congolese Swahili – French parallel text corpora)
as well as multiple corpora
from the OPUS collection \citep{tiedemann2012parallel}, namely:
ELRC\_2922 \citep{tiedemann2012parallel},
GNOME \citep{tiedemann2012parallel},
CCAligned and MultiCCAligned \citep{el2020ccaligned},
EUbookshop \citep{tiedemann2012parallel},
GlobalVoices \citep{tiedemann2012parallel},
JW300 for \texttt{sw} and \texttt{swc} source languages \citep{agic2019jw300},
ParaCrawl and MultiParaCrawl\footnote{\url{https://www.paracrawl.eu/}},
Tanzil \citep{tiedemann2012parallel},
TED2020 \citep{reimers2020making},
Ubuntu \citep{tiedemann2012parallel},
WikiMatrix \citep{schwenk2019wikimatrix} and
wikimedia \citep{tiedemann2012parallel}.
The validation data for each target language comprises of 434 randomly sampled
utterances from the shared task training data.
The testing data is the shared task validation data,
that also has 434 sentences per target language.

\begin{table}[ht]
  \centering
  \scriptsize
  \setlength{\tabcolsep}{3.5pt}
  \begin{tabular*}{0.92\columnwidth}{l|r|r|r|r}
    \noalign{\hrule height 1pt}
    Dataset & \multicolumn{2}{|c|}{Words} & \multicolumn{2}{|c}{Sentences} \\
    \cline{2-5}
     & \multicolumn{1}{|c|}{ \textrightarrow \texttt{eng}} & \multicolumn{1}{|c}{ \textrightarrow \texttt{fra}} & \multicolumn{1}{|c|}{ \textrightarrow \texttt{eng}} & \multicolumn{1}{|c}{\textrightarrow \texttt{fra}} \\
    \hline
     IWSLT'21          & 31,594 & 51,111 & 4,157     & 4,562 \\
     Gamayun           & 39,608 & 216,408 & 5,000     & 25,223 \\
     ELRC\_2922        & 12,691 & - & 607       & - \\
     GNOME             & 170 & 170 & 40        & 40 \\
     CCAligned         & 18,038,994 & - & 2,044,993 & - \\
     MultiCCAligned    & 18,039,148 & 10,713,654 & 2,044,991 & 1,071,168 \\
     EUbookshop        & 228 & 223 & 17        & 16 \\
     GlobalVoices      & 576,222 & 347,671 & 32,307    & 19,455 \\
     JW300 \texttt{sw} & 15,811,865 & 15,763,811 & 964,549   & 931,112 \\
     JW300 \texttt{swc}& 9,108,342 & 9,094,008 & 575,154   & 558,602 \\
     ParaCrawl         & 3,207,700 & - & 132,517   & - \\
     MultiParaCrawl    & - & 996,664 & -         & 50,954 \\
     Tanzil            & 1,734,247 & 117,975 & 138,253   & 10,258 \\
     TED2020           & 136,162 & 134,601 & 9,745     & 9,606 \\
     Ubuntu            & 2,655 & 189 & 986       & 53 \\
     WikiMatrix        & 923,898 & 271,673 & 51,387    & 19,909 \\
     wikimedia         & 66,704 & 1,431 & 771       & 13 \\
    \hline
    Total              & 41,910,113 & 30,003,158 & 3,406,772 & 2,159,007 \\
    \hline
    \noalign{\hrule height 1pt}
  \end{tabular*}
  \caption{
    \label{tab:data:mt} Datasets used to train the MT systems and their
    sizes in numbers of words (source language) and sentences.
    Total numbers are lower due to the deduplication.
  }
\end{table}

\begin{table*}[ht]
  \centering
  \small
  \begin{tabular*}{0.70\textwidth}{l|l|l|l|l|l|l|l}
    \noalign{\hrule height 1pt}
    Model & Input & \multicolumn{2}{|c|}{\texttt{swa}\textrightarrow \texttt{eng}} & \multicolumn{2}{|c|}{\texttt{swc}\textrightarrow \texttt{fra}} & \multicolumn{2}{|c}{\texttt{swc}\textrightarrow \texttt{eng}} \\
    \cline{3-8}
           &        & BLEU & chrF & BLEU & chrF & BLEU & chrF \\
    \hline
    \texttt{vanillaNMT}       & \texttt{textS}             & 25.72 & 53.47 & 17.70 & 44.80 & 10.55 & 38.07 \\
              & \texttt{asrS}       & 14.26 & 47.74 & 10.57 & 40.99 &  4.71 & 34.70 \\
              & ASR \#20        & 13.21 & 46.11 & 10.67 & 40.53 &  4.67 & 33.94 \\
              & ASR \#1         & 11.50 & 43.34 &  9.52 & 38.32 &  4.24 & 32.45 \\
    \hline
    \texttt{preprocNMT} & \texttt{textS}             & 11.01 & 41.33 & 13.54 & 41.00 &  4.49 & 31.91 \\
              & \texttt{asrS}       & 16.00 & 45.86 & 14.09 & 42.05 &  7.10 & 34.35 \\
              & ASR \#20        & 14.54 & 44.17 & 13.23 & 41.00 &  6.62 & 33.63 \\
              & ASR \#1         & 12.45 & 40.95 & 11.21 & 38.08 &  5.47 & 31.63 \\
    \hline
    \texttt{multiNMT} & \texttt{textS}             & 25.69 & 53.27 & 18.20 & 44.66 & 10.56 & 38.29 \\
                 & \texttt{asrS}       & 20.07 & 50.31 & 14.69 & 43.07 &  8.73 & 36.72 \\
                 & ASR \#20        & 17.91 & 48.39 & 13.29 & 41.58 &  7.94 & 35.47 \\
                 & ASR \#1         & 15.81 & 45.31 & 11.97 & 39.09 &  7.03 & 33.78 \\
    \noalign{\hrule height 1pt}
  \end{tabular*}
  \caption{
    \label{tab:results:mt} MT results on the shared task validation data.
    WER values on \texttt{swa}/\texttt{swc} validation data are 13.6/18.7\% for ASR \#20 and 25.9/26.5\% for ASR \#1.
  }
\end{table*}

\subsection{Model}
\label{subsec:mtmodel}
For the text-to-text neural machine translation (NMT) system we use a Transformer big model \citep{vaswani2017attention} using the fairseq implementation \cite{ott-etal-2019-fairseq}. We train three versions of the translation model.

First we train a vanilla NMT (\texttt{vanillaNMT}) system using only the data from the parallel training dataset. For preprocessing we use the SentencePiece implementation \cite{kudo2018sentencepiece} of BPEs \cite{sennrich2016neural}. For our second experiment for the NMT system (\texttt{preprocNMT}), we apply the same written to spoken language conversion as used for the ASR transcriptions (section \S\ref{subsec:asrdata}) to the source text $S$ and obtain ASR-like text $S_t$. $S_t$ is then segmented using a BPE model and used as input for our NMT model. The last approach was using a multi-task framework to train the system (\texttt{multiNMT}), where all parameters of the translation model were shared.  The main task of this model is to translate ASR output $S_t$ to the target language $T$ (task \texttt{asrS}), while our auxiliary task is to translate regular source Swahili $S$ to the target language $T$ (task \texttt{textS}). We base or multi-task approach on the idea of multilingual NMT introduced by \newcite{johnson-etal-2017-googles}, using a special token at the beginning of each sentence belonging to a certain task, as we can see in the next example:

\vspace{4pt}
\noindent
\texttt{<asrS>} sara je haujui tena thamani ya kikombe hiki $\to$  Tu ne connais donc pas, Sarah, la valeur de cette coupe ?

\noindent
\texttt{<textS>} Sara, je! Haujui tena, thamani ya kikombe hiki? $\to$ Tu ne connais donc pas, Sarah, la valeur de cette coupe ?
\vspace{4pt}

Then, our multi-task training objective is to maximize the joint log-likelihood of the auxiliary task \texttt{textS} and the primary task \texttt{asrS}. 

\paragraph{Hyperparameters} For word segmentation we use BPEs \cite{sennrich2016neural} with separate dictionaries for the encoder and the encoder, using the SentencePiece implementation \cite{kudo2018sentencepiece}. Both vocabularies have a size of 8000 tokens. Our model has 6 layers, 4 attention heads and embedding size of 512 for the encoder and the decoder. To optimize our model we use Adam \cite{kingma2014adam} with a learning rate of 0.001. Training was performed on 40 epochs with early stopping and a warm-up phase of 4000 updates. We also use a dropout \cite{JMLR:v15:srivastava14a} of 0.4, and an attention dropout of 0.2. For decoding we use Beam Search, with a size of 5.

\subsection{Results}
\label{subsec:mtresults}

Table \ref{tab:results:mt} shows the results of our MT system in combination with different inputs. We trained three models using the techniques described in section \S\ref{subsec:mtmodel} (\texttt{vanillaNMT, preprocNMT}, and \texttt{multiNMT}). Then we used the official validation set as input (\texttt{textS}), and also applied \texttt{asrS} preprocessing. We used both inputs to test the performance of all models with different inputs. As expected, the \texttt{vanillaNMT} systems performs well with \texttt{textS} input (i.e 25.72 BLEU for \texttt{swa}\textrightarrow \texttt{eng}), but drops when using \texttt{asrS}. This pattern was later confirmed when using real ASR output (ASR \#20 and ASR \#1). We noticed, that training our model with \texttt{asrS}, instead of using \texttt{textS} improves slightly the results (i.e 16.00 BLEU with \texttt{preprocNMT} compared with 14.26 on \texttt{vanillaNMT} for \texttt{swa}\textrightarrow \texttt{eng}). But when we use \texttt{multiNMT} the performance strongly increase to 20.07 for \texttt{swa}\textrightarrow \texttt{eng}. This pattern also can be seen when using real ASR output (ASR \#20 and ASR \#1), and across all language pairs. We hypothesize  that the multi-task framework helps the model to be more robust to different input formats, and allows it to generalize more the language internals.

\section{End-to-End ST}
\label{sec:endtoend}

\subsection{Data}
\label{subsec:endtoenddata}

End-to-end ST is fine-tuned on the same speech recordings,
as ASR data, but with transcriptions in English or in French. English and French
transcriptions are obtained either from the datasets released with the shared task,
or by running our MT system on Swahili transcriptions.
External LMs for English and French outputs are trained on 10M sentences
of the corresponding language from the
OSCAR corpus \citep{ortiz-suarez-etal-2020-monolingual}.

\subsection{Model}
\label{subsec:endtoendmodel}

The end-to-end ST system comprises of the Encoder part of our ASR system
and the whole MT system with removed input token embedding layer.
All layers are frozen during the fine-tuning except of the top four
layers of ASR Encoder and bottom three layers of MT Encoder.
SpecAugment and gradient accumulation are disabled during the fine-tuning.
Compared to the ASR system, end-to-end ST system has larger dictionary,
what leads to shorter output sequences and allows us to increase the batch size
to 60M bins.
The rest of hyperparameters are the same as in the ASR system.
We evaluate ST model separately and also with external LM that is set up
as described in the ASR section.

\subsection{Results}
\label{subsec:endtoendresults}

It can be seen from Table \ref{tab:results:st} that the end-to-end ST systems do not yet
match the cascaded systems in translation quality in low resource settings.
External LMs, however, slightly improve the results for both target languages.

\begin{table}[ht]
  \centering
  \small
  \setlength{\tabcolsep}{2.5pt}
  \begin{tabular*}{1.02\columnwidth}{l|l|l|l|l|l|l}
    \noalign{\hrule height 1pt}
    Setting & \multicolumn{2}{|c|}{\texttt{swa}\textrightarrow \texttt{eng}} & \multicolumn{2}{|c|}{\texttt{swc}\textrightarrow \texttt{fra}} & \multicolumn{2}{|c}{\texttt{swc}\textrightarrow \texttt{eng}} \\
    \cline{2-7}
                        & BLEU  & chrF  & BLEU  & chrF  & BLEU & chrF \\
    \hline
    No LM               & 7.81  & 30.83 & 2.94  & 22.98 & 3.59 & 23.50 \\
    LM weight 0.4       & 8.82  & 31.26 & 3.73  & 22.07 & 4.06 & 23.89 \\
    LM weight 0.6       & 9.11  & 31.45 & 3.58  & 20.57 & 4.17 & 23.62 \\
    \hline
    \noalign{\hrule height 1pt}
  \end{tabular*}
  \caption{
    \label{tab:results:st} End-to-end ST results on the shared task validation data.
  }
\end{table}

\section{Final systems}
\label{sec:finalsystem}

Table \ref{tab:results:final} shows validation scores of our final systems,
as well as their evaluation scores provided by the organizers of the shared task.
Our primary (cascaded) system here uses increased beam sizes: 30 for the ASR, 10 for the English MT
and 25 for the French MT. \texttt{swc/swa} WERs of the final ASR systems are 12.5/17.6\% on the validation sets.
We did not observe improvement from the increased beam size on the contrastive systems
and leave it at 2.
It should be noted that the contrastive system is evaluated on
incomplete output\footnote{406 of 2124 hypothesis are empty.}
for the \texttt{swc}\textrightarrow \texttt{fra} pair because of the technical issue on our side.
We observe a large gap between the validation and evaluation scores for Coastal Swahili source
language, what might indicate some sort of bias towards the validation set in our ASR or MT, or both.
It is unclear why it does not happen for Congolese Swahili source language, because
we optimized all our systems for the best performance on the validation sets for both source languages.

\begin{table}[ht]
  \centering
  \small
  \setlength{\tabcolsep}{2.5pt}
  \begin{tabular*}{0.98\columnwidth}{l|l|l|l|l}
    \noalign{\hrule height 1pt}
    System & Set & \multicolumn{1}{|c|}{\texttt{swa}\textrightarrow \texttt{eng}} & \multicolumn{1}{|c|}{\texttt{swc}\textrightarrow \texttt{fra}} & \multicolumn{1}{|c}{\texttt{swc}\textrightarrow \texttt{eng}} \\
    \hline
    Primary & Val.       & 18.3 & 13.7 & 7.9 \\
    \cline{2-5}
    (cascaded)  & Eval.       & 14.9 & 13.5 & 7.7 \\
    \hline
    Contrastive & Val.       & 9.1 & 3.7   & 4.0 \\
    \cline{2-5}
    (end-to-end) & Eval.       & 6.7 & 2.7   & 3.9 \\
    \noalign{\hrule height 1pt}
  \end{tabular*}
  \caption{
    \label{tab:results:final}
    Results (BLEU) of the primary and contrastive systems on the validation and evaluation data of the shared task.
  }
\end{table}

\section{Conclusion}
\label{sec:conclusion}

This paper described the IMS submission to the IWSLT 2021 Low-Resource Shared Task on Coastal and Congolese Swahili to English and French, explaining our intermediate ideas and results. Our system is ranked as the best for Congolese Swahili to French and English, and the second for Coastal Swahili to English. In spite of the simplicity of our cascade system, we show
that the improving of ASR system with pre-trained models and afterward the tuning of MT system to optimize its fit to the ASR output achieves good results, even in challenging low resource settings. Additionally, we tried an end-to-end ST system with a lower performance. However, we learned that there is still room for improvement, and in future work we plan to investigate this research direction.

\section{Acknowledgements}
\label{sec:acknowledgements}
The authors acknowledge support by the state of Baden-Württemberg through bwHPC.
This product contains or makes use of IARPA data,
IARPA Babel Swahili Language Pack, IARPA-babel202b-v.1.0d (LDC2017S05).
Manuel Mager received financial support
by DAAD Doctoral Research Grant for this work.

\bibliographystyle{acl_natbib}
\bibliography{iwslt2021}


\end{document}